\documentclass[conference]{IEEEtran}
\IEEEoverridecommandlockouts
\usepackage{amsmath,amssymb,amsfonts}
\usepackage{algorithmic}
\usepackage{graphicx}
\usepackage{textcomp}
\usepackage{xcolor}
\def\BibTeX{{\rm B\kern-.05em{\sc i\kern-.025em b}\kern-.08em
    T\kern-.1667em\lower.7ex\hbox{E}\kern-.125emX}}

\usepackage{microtype}
\usepackage{graphicx}
\usepackage{subfigure}
\usepackage{booktabs} %
\usepackage{amsmath}
\usepackage{amssymb}
\usepackage{multirow}
\usepackage{hyperref}
\usepackage[numbers]{natbib}

\newcommand{\xhdr}[1]{\vspace{4pt}\noindent{\textbf{#1}}}
\newcommand{\eg}[0]{\textit{e.g., }}
\newcommand{\ie}[0]{\textit{i.e., }}

\begin{document}

\title{Action Recognition based on Cross-Situational Action-object  Statistics}

\author{}

\author{
  \IEEEauthorblockN{Satoshi Tsutsui\IEEEauthorrefmark{1},
    Xizi Wang\IEEEauthorrefmark{2},
    Guangyuan Weng\IEEEauthorrefmark{3},
    Yayun Zhang\IEEEauthorrefmark{4},
    David J. Crandall\IEEEauthorrefmark{2}, and
    Chen Yu\IEEEauthorrefmark{4}}
  \IEEEauthorblockA{\IEEEauthorrefmark{1}National University of Singapore, Singapore,    satoshi@nus.edu.sg}
  \IEEEauthorblockA{\IEEEauthorrefmark{2}Indiana University, Bloomington, IN USA,
    \{xiziwang,djcran\}@iu.edu}
  \IEEEauthorblockA{\IEEEauthorrefmark{3}Northeastern University, Boston, MA USA,
    weng.g@northeastern.edu
  }
  \IEEEauthorblockA{\IEEEauthorrefmark{4}University of Texas at Austin, Austin, TX USA,
    \{yayunzhang,chen.yu\}@utexas.edu}
  }

\maketitle

\begin{abstract}
Machine learning models of visual action recognition are typically
trained and tested on data from specific situations where actions are
associated with certain objects. It is an open question how
action-object associations in the training set influence a model's
ability to generalize beyond trained situations. We set out to
identify properties of training data that lead to action recognition
models with greater generalization ability.  To do this, we take
inspiration from a cognitive mechanism called cross-situational
learning, which states that human learners extract the meaning of
concepts by observing instances of the same concept across different
situations. We perform controlled experiments with various types of
action-object associations, and identify key properties of
action-object co-occurrence in training data that lead to better 
classifiers. Given that these properties are missing in the datasets
that are typically used to train action classifiers in the computer
vision literature, our work provides useful insights on how we should
best construct datasets for efficiently training for better
generalization.
\end{abstract}

\section{Introduction}

\begin{figure}[t]
    \centering
	\includegraphics[width=\linewidth]{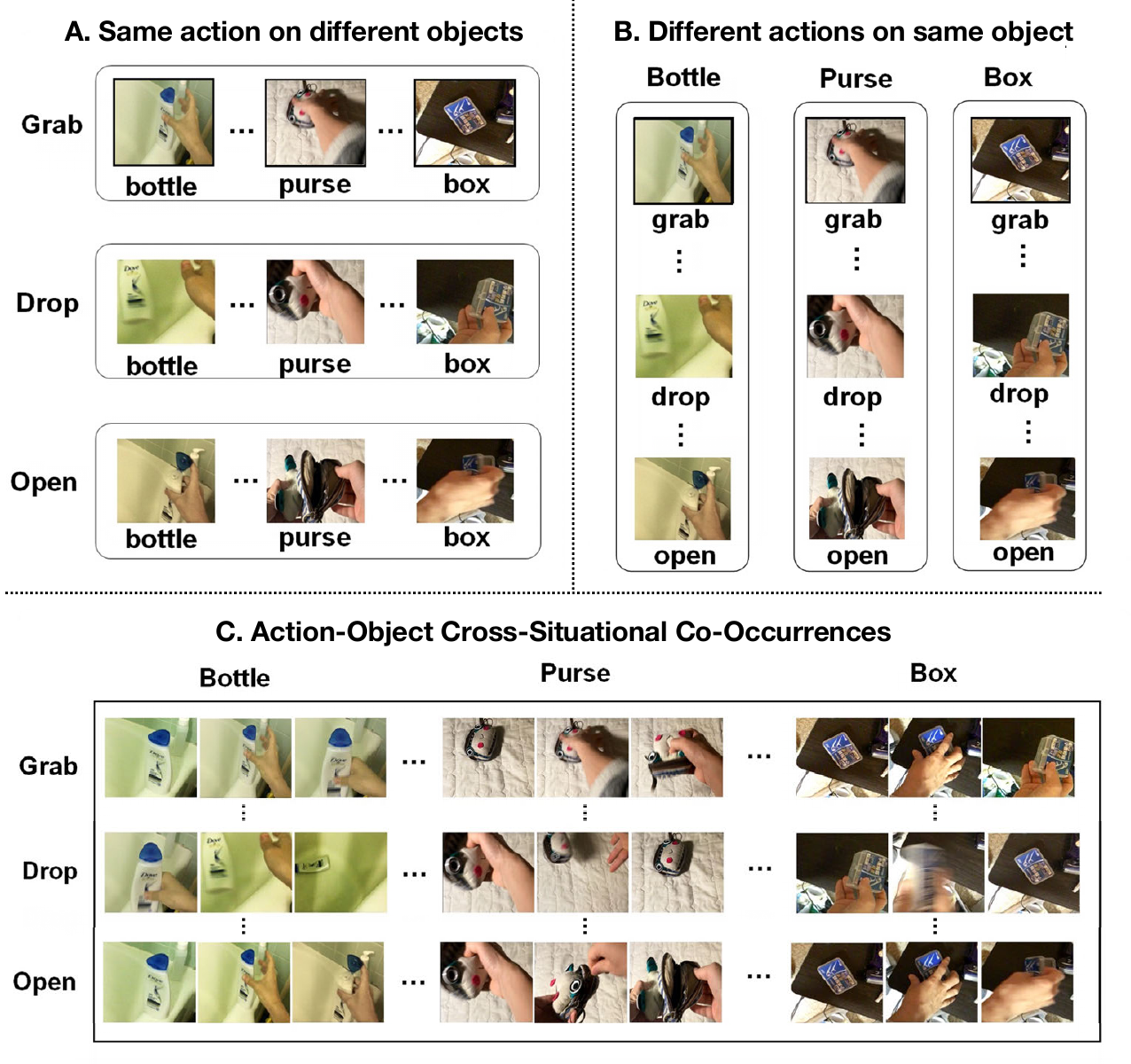}
  \caption{ We take inspiration from the cognitive mechanism of
    cross-situational learning to investigate how to best design
    action recognition training data.
    (A) Each action is applied to multiple objects, so cross-situational learning can identifying common elements belonging to the action.
    (B) Each action is applied to 
    the same object, so cross-situational learning
    can learn to ignore the identity of the object and focus
    on properties of different actions. (C) These two kinds of
    cross-situational statistics are represented in an
    action-object matrix. 
    The challenge in action recognition is to
    efficiently train models 
    from these action-object
    co-occurrences so 
    that  it can generalize to actions 
    on objects
    not seen in training. 
    }
\label{fig:intro}
\end{figure}

Teaching computer vision systems to recognize actions poses a challenging generalization problem because the same action can be applied to many target objects, including those not in the training dataset.  For example, the top row
of Figure~\ref{fig:intro}A shows three instances of the ``grab''
action. If trained on these instances, the learning system needs to
learn that ``grab'' is the action applied to an object, not a property
of any of these specific objects (bottle, purse, box) themselves.
Generalization in action recognition models requires recognizing all
instances of an action in the real world, including on target objects
not seen during training.

While many action recognition techniques have been proposed in the
computer vision literature, the vast majority of this work introduces
new algorithms or neural network models.  However, a fundamental open
question that has not yet been explored in depth is how the structural
properties of training datasets affect the accuracy and generalization
of action recognition models. Most existing action recognition
datasets have very limited or specific action categories, each
associated with very few unique objects. For example,
Kinetics-700~\cite{kineticscarreira2017quo} includes the
highly-specific \textit{picking blueberries} action; to recognize this
action, the model could simply learn to detect blueberries instead of
a general representation of the ``picking'' action.  In the current
study, our goal is to explore how the \textit{structure} of the
training dataset affects a model's ability to learn generalizable
representations of actions, which could help 
to better design training data collection when developing action recognition
systems. Towards this end, we created training datasets containing
different action-object co-occurrence statistics and compared model
performances.

To create such training datasets, we take inspiration from cognitive
science and developmental psychology. Cross-situational
learning~\cite{yu2007rapid} is a general learning mechanism suggesting
that human learners can learn a concept better if they are presented
with instances of the concept in many different conditions. Imagine
how a young child learns the meaning of a new object that she has
never seen before --- ``ball,'' for example. After seeing the object
in different contexts while hearing the word ``ball,'' the child could
discover that while the word is heard when many other objects are also
present, it consistently accompanies a round, bouncy object.

This
cognitive ability to use co-occurrence statistics to discover reliable
patterns across contexts can be applied to learning more complicated
concepts such as actions representing relations between objects. For
example, if the learner hears ``cutting'' while viewing multiple
instances of cutting apples, it is unclear if the word refers to the
action or the object, since both appear in all learning instances.
But if the learner hears the word ``cutting'' while viewing instances
of cutting meat, cutting apples, and cutting sheets of paper, it may be easier
to understand that the action of cutting is the common element, and
that the specific objects (meat, apple, paper) are not important and
should be ignored \cite{liu2019verb,zhang2020verb}.

Inspired by the principles of cross-situational learning demonstrated
in human learners \cite{yu2007rapid}, this paper investigates how to
effectively structure action-object cross-situational statistics so
that machine learning models can better recognize actions and make
generalizations. In the real world, some objects are often acted on by many
possible actions while others are not. For example, many ``common''
objects (\eg \textit{box, chopsticks, beef,} etc.) can be the target of the
 action \textit{grasp}, but only some ``unique'' objects 
(\eg \textit{beef, chicken, mushrooms}, etc.) can be the target of the
action \textit{roast}. Based on these observations, two types of
co-occurrence statistics were created in the current study: (1)
same-action-to-different-objects (Figure~\ref{fig:intro}A) and (2)
different-actions-to-same-object (Figure~\ref{fig:intro}B). To study
how the two types of action-object statistics affect recognition
performance, we systematically created multiple training datasets
containing different combinations of the common and unique objects for
each action. We trained multiple state-of-the-art action classifiers
and compared classification accuracy across different training
datasets and across different models.

In contrast to previous studies reporting overall accuracy, the
experiments in this paper were conducted in a much more rigorous way
by dividing testing video instances into three categories: (1)
generalization to instances containing common objects in the training
data, (2) generalization to instances containing unique objects in the
training data, and (3) generalization to instances containing unseen
objects which never appeared in the training data. This systematic
testing method allows us to understand the generalization ability of
the models in a fine-grained manner.

\xhdr{Contributions.} In short, our paper has three main
contributions:

\begin{enumerate} \item To the best of our knowledge, this study
is the first to introduce the human learning principle of
cross-situational learning to the machine learning community for
addressing action recognition. \item We perform systematic
experiments on both training and testing data to theoretically
investigate the effects of different action-object co-occurrences on
generalization. \item We demonstrate that having more common
objects across different actions in the training data improves
recognition performance for unseen action-object instances, while
having objects uniquely tied with an action has only limited
contribution to generalization.
\end{enumerate}
Taken together, these contributions may provide practical
guidelines on how to 
to better construct efficient training data when developing future action
recognition systems.

\section{Related Work}
Broadly speaking, our work is related to \textit{affordance}, which
refers to the property of  an object that elicits human
actions~\cite{jamone2016affordances}. Our work is also related to
cross-situational learning, and many studies~\cite{taniguchi2017cross,
  juven2020cross, chen2016experimental} on it have been conducted in
the field of developmental learning and robotics. Our study also builds on other work that has
used insights from developmental psychology to try to improve computer vision.
For example, several
studies~\cite{Bambach2018,orhan2020self,bambach2017egocentric}
have
analyzed videos collected from cameras mounted on heads of toddlers, and have demonstrated that 
children's visual perspectives have important structural properties that may help inform how to better train computer vision
models of object recognition. We have similar motivation to them, and
apply cross-situational learning into the design of training data
for computer vision models of action recognition.

\begin{figure*}
\centering
	\includegraphics[width=0.85\linewidth]{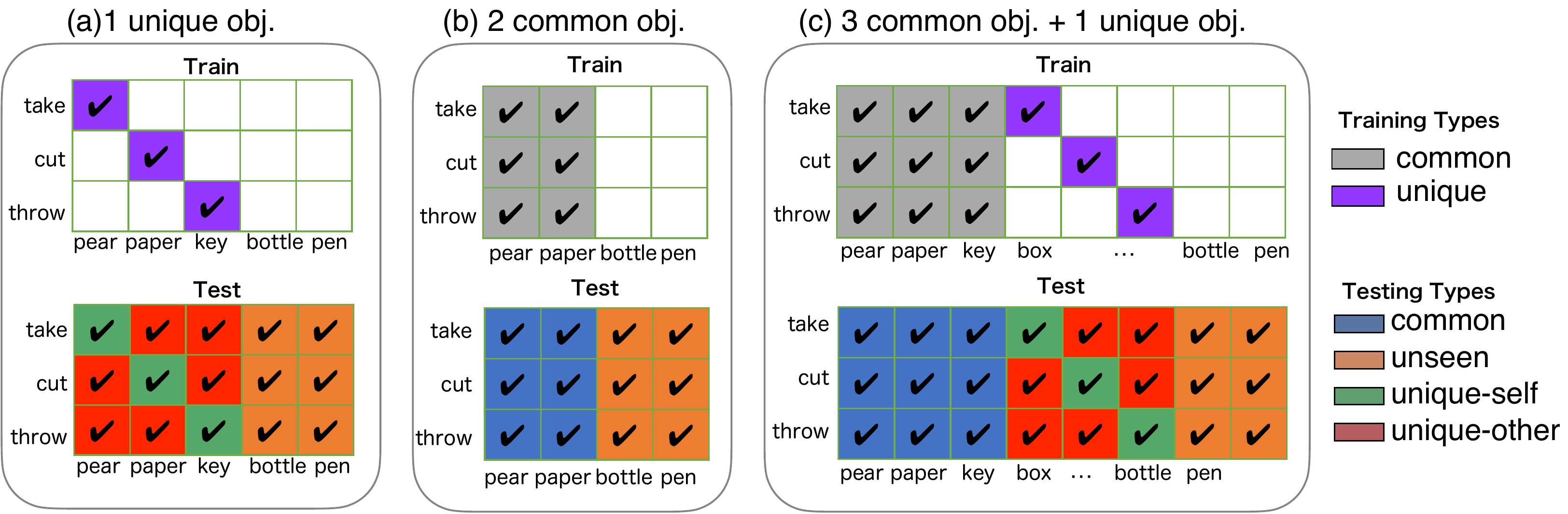}
  \caption{Modeling the essence of training/testing co-occurrences for action recognition. See Table~\ref{tbl:object-types} for more detail.}
\label{fig:train-test}
\end{figure*}

\begin{table*}[t]
    \caption{Terminology we use for our action-object statistics in training and testing.}
    \label{tbl:object-types}
\centering
    \begin{tabular}{|c|l|l|l|}
    \hline \multirow{2}{*}{train} & \multicolumn{2}{l|}{common} & instances with objects that are used in training and they co-occur with all of the actions. \\ \cline{2-4} & \multicolumn{2}{l|}{unique} & instances are used in training, and they only co-occur with a particular action.  \\ \hline \multirow{4}{*}{test} & \multicolumn{2}{l|}{common} & instances contain common objects and will be used at test.  \\ \cline{2-4} & \multirow{2}{*}{unique} & self & instances contain unique objects for that action.  \\ \cline{3-4} & & other & instances contain unique objects for other actions.  \\ \cline{2-4} & \multicolumn{2}{l|}{unseen} & instances with objects that are never used in training with any associated action.  \\ \hline
    \end{tabular}
\end{table*}

\section{Methodology}

Many action recognition datasets in computer vision (e.g., UCF101~\cite{soomro2012ucf101} and
Kinetics~\cite{kineticscarreira2017quo}) treat action-object pairs as action
classes, such that the concepts of object and
actions are entangled. For example, Kinetics includes ``cleaning toilet'' and ``cleaning windows'' as two
distinct action classes, and no other action classes include the ``toilet'' or ``window'' objects. 
Models trained on such a dataset would find it
nearly impossible to generalize to ``cleaning'' on novel objects,
because the learning problem can be solved simply by recognizing the presence of toilets or windows.

We argue
that actions and objects need to be treated as different sets of
labels in order to learn abstract knowledge
from their co-occurrence.  Epic
Kitchens~\cite{Damen_2018_ECCV} and Something-Something~\cite{goyal2017something} annotate actions and objects separately. Epic Kitchens is biased to kitchen environments while Something-Something~\cite{goyal2017something} is more general, so
we leverage Something-Something  for our experiments.

This section explains how we model and manipulate action-object
co-occurrences for training and testing action classifiers.  We
investigate two primary types of structure while training action
classifiers, and four types of testing, as we summarize in
Table~\ref{tbl:object-types} and Figure~\ref{fig:train-test} and describe in the next sections.  These
experimental settings allow us to apply the cross-situational learning
in a principled manner.

\subsection{Action-object co-occurrence for training}  While the
real world has complex interactions between actions and objects, we believe
the key elements of action-object co-occurrence for training action
classifiers lie in two extreme types of interaction -- \textit{unique}
and \textit{common}. For example, in a kitchen environment,
\textit{put} is \textit{common} because it can be used on almost any 
object (carrot, paper, spoon, etc.), whereas \textit{cut} is
\textit{unique} because it is mostly only used with some objects (carrot, cabbage,
apples, etc).

To illustrate this, let us think about a simplified world with only three
actions (\textit{take}, \textit{cut}, and \textit{throw}), and
consider Figure~\ref{fig:train-test}-(a)-Train. Each action has a
single and distinct object category associated with it; we refer to
these as \textit{unique} objects. In the example, \textit{pear},
\textit{paper}, and \textit{key} are the unique object of
\textit{take}, \textit{cut}, and \textit{throw}, respectively. In Figure~\ref{fig:train-test}-(b)-Train, both \textit{pear}
and \textit{paper} are used with all three actions, so these are
\textit{common} objects. Lastly, training data can have both
\textit{common} and \textit{unique} objects, as in
Figure~\ref{fig:train-test}-(c)-Train.

\subsection{Action-object co-occurrence for testing} After training action
classifiers with datasets that consist of examples with combinations of common and
unique objects, we need to evaluate the performance of the models using
some held-out instances of action-object pairs.
When evaluating, we identify each testing action-object pair
as one of four types with respect to the training data:
\begin{itemize}
\item \textbf{Common}
testing objects are those whose category is \textit{common} in the
training data.
\item \textbf{Unseen} testing objects are never
used in the training, and are important for measuring the generalization
ability of the model.
\item \textbf{Unique-self} testing objects are those for which the
object category is \textit{unique} in the training set and the action-object pair was seen in training.
\item \textbf{Unique-other} testing objects, which are also critical to
test the generalization ability of the models, are those for which the
object category is \textit{unique} in the training set but the action-object pair was not seen in training.
\end{itemize}

As an example, suppose that our training data consists only of
\textit{cut-onions} and \textit{roast-chicken} instances.  There are no
possible \textbf{common} testing objects in this example since their
are no common objects in the training set.  \textbf{Unseen}
action-object testing pairs could include \textit{cut-apples} or
\textit{roast-duck}, since these objects were not seen in the training
set. \textbf{Unique-self} testing objects would include when 
\textit{onion} is evaluated with \textit{cut} and \textit{chicken}
with \textit{roast}, since these pairs appeared in training.
\textbf{Unique-other} would include when the object \textit{onion} is tested with
\textit{roast} (and \textit{chicken} with \textit{cut}), since these instances never
appeared in training.

\subsection{Manipulations and Assumptions} Of course, these are simplified
models of the complex action-object statistics in the real world. However,
for the purpose of systematically studying the effect of action-object
structure in training data, we believe that the above abstraction
captures the essence of the co-occurrences. We can then manipulate the
number of \textit{unique} and \textit{common} objects
in the training set, and evaluate on the four types of objects
(\textit{common}, \textit{unique-self}, \textit{unique-others}, and
\textit{unseen}).  These manipulation patterns are
carefully designed so that we can meaningfully compare performances of
the trained models.

Importantly, our aim is to identify general
principles that are not dependent on specific actions or objects.
We
therefore make a number of  assumptions.  First, we use actions
with the same level of concreteness so that we can factor out 
inherent differences between actions.  For example,
\textit{repair} is less concrete than
 \textit{put} or
\textit{grasp} and involves many different sub-actions (potentially including
\textit{put} and \textit{grasp}), so we do not use it.  Second, for
each experiment, we arrange so that all actions have the same number of common objects
and the same number of unique objects. Third, we assume that instances
of the same object category have relatively little within-class
variation and that each object instance equally contributes to the
action classifier if it is added to the training set.

\begin{figure*}[htb!]
\centering
	\includegraphics[width=\linewidth]{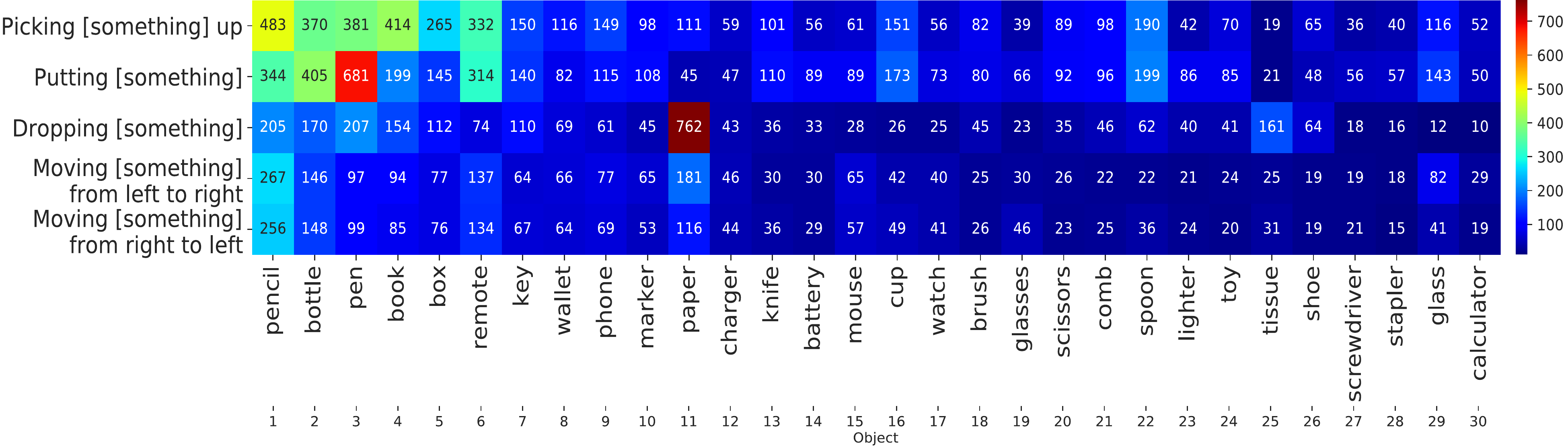}
  \caption{Co-occurrence matrix showing frequencies of action-object instances in our dataset, which we sub-sampled from the Something-Something dataset.}
\label{fig:matrix-something-selected}
\end{figure*}

\section{Experiments}
\subsection{Setup}
To conduct experiments into the cross-situational learning of action recognition
models, we need a dataset that has dense action-object co-occurrences --- where actions are not tied
to specific objects, and there is a wide variety of target objects for each action. Since
most existing datasets have specific action-object
ties or very sparse co-occurrence matrices, so we had to tailor them
for our needs. We use
Something-Something~\cite{goyal2017something} because it includes a
wide variety of action-object pairs in many contexts. However, the
dataset has more than 100 action categories with different levels of
granularity, and the co-occurrence matrix is quite sparse. Hence, we
selected a subset of actions and objects in order to facilitate our
controlled experimentation.

 For making the dense subset, we started with the 10 merged action
 classes provided by the dataset. We sub-sampled the objects and
 actions by using a greedy algorithm with a minimum threshold on the
 frequency and the proportion of non-zero elements per row and per
 column of the co-occurrence
 matrix. Figure~\ref{fig:matrix-something-selected} shows a
 sub-sampled co-occurrence matrix in which 5 actions and 30 objects
 are selected.  Note that all cells are filled with at least 10
 instances. We keep objects \#21 to \#30 as unseen objects for
 testing, and sample common and unique objects from the rest for
 training. We bias the sampling in order to obtain approximately the
 same number of instances for each non-zero cell in the training
 matrix, to try to mitigate the effect of the class imbalance.  Unless
 otherwise noted, we sample a total number of 375 instances (i.e., we
 have 375 training samples by default) for each training matrix, which
 means 75 instances per action class.  We do the sampling ten times,
in each case training an action classifier, and report the mean accuracy and 95\%
 confidence intervals. Our default action classifier is a
 3D-ResNet18~\cite{hara2018can}, which takes a video clip as input
 and outputs  one of the five actions selected above. We use
 PyTorch's default 3D-ResNet18 implementation and make the source code
 available.\footnote{\url{http://vision.soic.indiana.edu/cross-situational-action-recognition/}}

 Throughout the following sections, we report accuracies of multi-way action classification problems. We do not report
 object recognition results because that is not our goal: our goal is to recognize an action regardless of whether
 the accompanying object was used for training or not.

\subsection{Effects of action-object statistics}\label{sec:main-exp}
We first investigate the effects of training with either unique objects only or common
objects only. We also apply combinations of both unique
and common objects to investigate the synergistic effects between them.

\subsubsection{Unique Objects}\label{results:unique}
\begin{figure}
    \centering
    	\includegraphics[width=\linewidth]{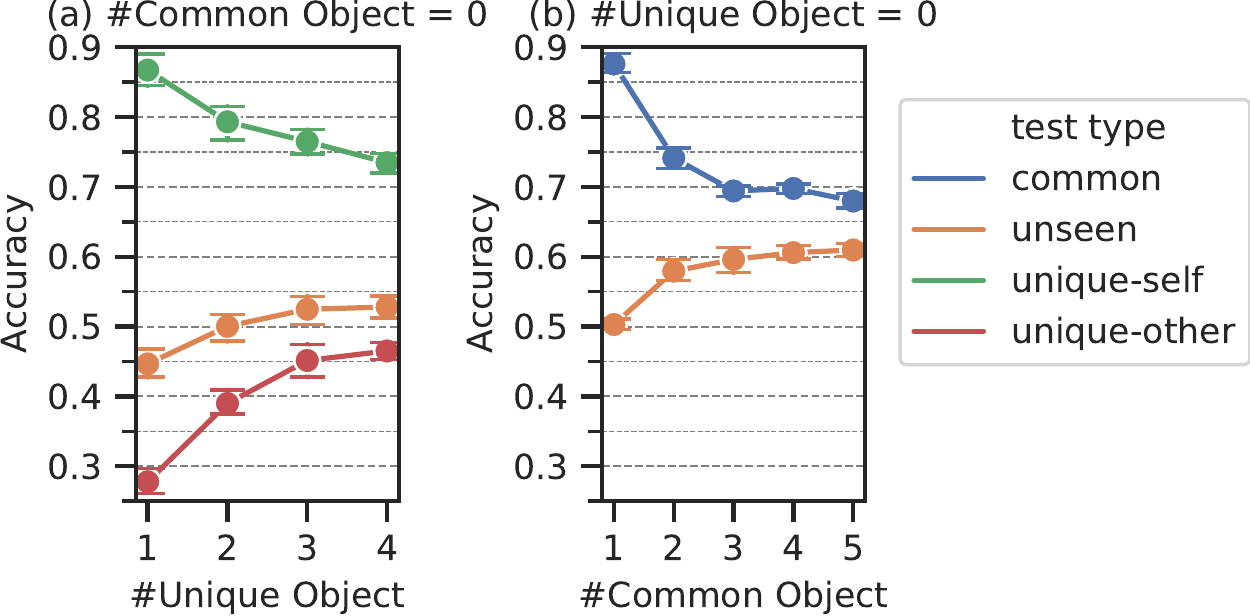}
     \caption{(a) Accuracies of three testing conditions, \ie unseen, unique-self, and unique-other given that the number of common objects is 0, while increasing the number of unique objects. The plot shows that increasing unique objects does not really help for generalization. (b) Accuracies of two testing conditions, \ie common and unseen, given number of unique objects is 0, while increasing the number of common objects. The plot shows that a greater number of common objects results in better unseen accuracy. } %
    \label{fig:exp-unique-only} \label{fig:exp-common-only}
\end{figure}

We gradually increase the number of unique objects from one to four and observe the effect on model performance.
Figure~\ref{fig:exp-unique-only}-(a) presents the results, showing
accuracies of three testing conditions (unseen, unique-self, and unique-other).  Regardless of the number of
unique objects, the accuracies for the unique-self objects are always
high, the accuracies of unseen cases are lower, and unique-others
are the lowest. This suggests that the model mostly memorizes the
objects instead of actually learning the essential properties of the
actions themselves: the model can classify the action very well only if the
same action-object was seen in training samples,
which is a severe drawback in practice.  Increasing
the number of unique objects improves the performance for unseen and
unique-other cases, and decreases the performance for unique-self
objects. The reason for the decrease could be that as the number of unique objects increases,
the model has to
memorize more objects under the same action category, although simply remembering
objects is not a desirable behavior.

\begin{figure*}
\centering
	\includegraphics[width=0.9\linewidth]{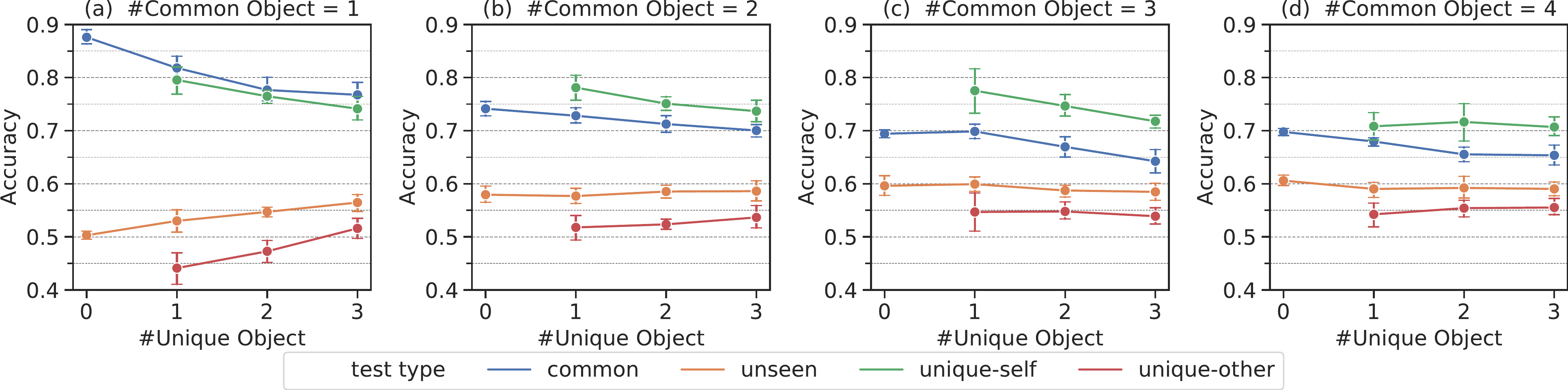}
  \caption{Different combinations of common and unique objects. Adding common unique objects has limited effect on generalization. }
\label{fig:exp-common-unique}
\end{figure*}

\subsubsection{Common Objects}
  Figure~\ref{fig:exp-common-only}-(b) shows the results of 1 to 5
  common objects. The test accuracy on common objects is always higher
  than on unseen objects.  This shows that the model can better
  recognize the action if it has seen instances of the object
  associated with the action. The accuracy for testing common objects
  decreases as we increase the number of training common objects,
  because the model must remember more action-object combinations,
  making the task harder.

In contrast, the accuracy on unseen objects increases up to three and
then reaches a plateau. This suggests that having more common objects in
the training data increases the model's generalization ability. We
conjecture two reasons for the plateau at three: (1) the
task is to choose one of just five actions, so three common objects
may already be sufficiently cross-situational for the model, or  (2) we use the
object category as a single concept but the visual variation inside
each category is high enough so adding a single category contributes
much more diversity than we might expect.

\subsubsection{Common and Unique Objects}
How about the comparison between common and unique objects? We can compare
the unseen object accuracy to check this. The models trained from
unique objects (Figure~\ref{fig:exp-common-only}-(a)) have lower unseen
accuracy than the models trained from common objects
(Figure~\ref{fig:exp-common-only}-(b)).  This indicates that common
objects are more helpful for capturing the meaning of actions than
unique objects.  Then, how about the synergy between them? If we add some
unique objects to the common objects, would they introduce additional
diversity into the training set and help build stronger models?  To
answer this, we added one to three unique objects into each case of
common objects only. The results are shown in
Figure~\ref{fig:exp-common-unique}. Except for the case of the single
common object, adding unique objects did not affect the
results. However, interestingly, adding unique objects slightly
improved the accuracy of unseen and unique-others for the case of a
single common object.  This suggests that adding unique objects is
helpful if it is used for the case of very limited common objects.

\subsection{Effect of Quantity}
\begin{figure}
\centering
	\includegraphics[width=0.7\linewidth]{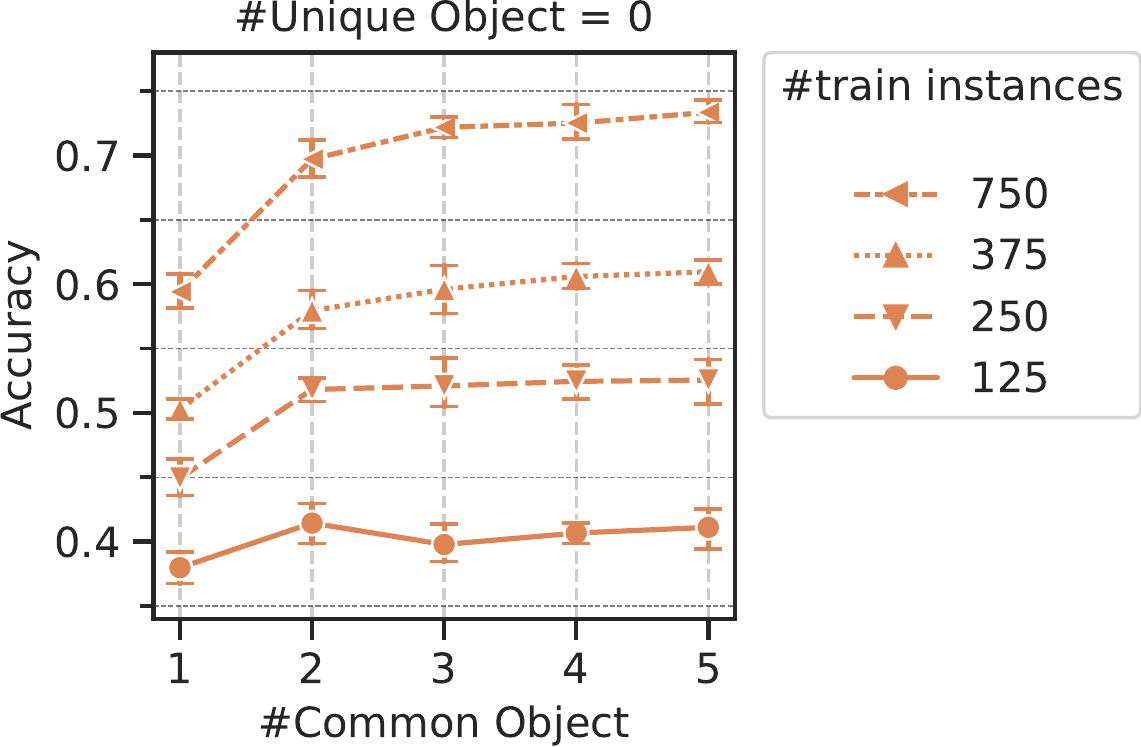}
  \caption{Accuracy tested on unseen object categories improves as we increase the number of common objects for various size of training datasets.  }
\label{fig:exp-num}
\end{figure}

We have used the same total number of instances for all the
experiments so far because we wanted to study the effects
of training data quality rather than quantity.  However, deep learning
generally benefits from having as many training examples as possible,
so we also experimented with different numbers of total instances.
Given that unique objects are not very helpful, we investigated the
effect of quantity by re-running the common object experiments with
different numbers of total instances, and evaluating the unseen accuracy,
which measures generalization.  Figure~\ref{fig:exp-num} shows the
results. As we expected, larger training data sizes indeed improve
performance.  However, sometimes the effect of having common objects
is more important than quantity: for example, the accuracy of four
common objects with a total of 375 examples is higher than a single
common object with a total of 750 examples.  Overall, except for the case of 125 instances, we observe the same trend that more common objects
improves accuracy.  For the case of 125, presumably because the total
number of instances is too low, controlling the quality does not
really matter.

\begin{figure*}
\centering
	\includegraphics[width=0.9\linewidth]{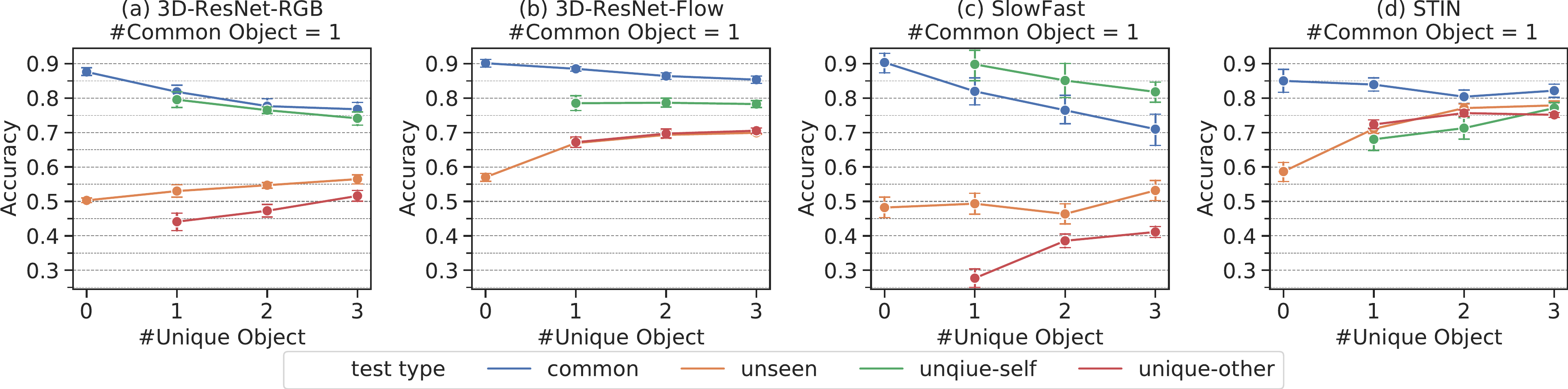}
  \caption{Test accuracies over four different models. Overall we observe relatively similar trends except for STIN, which turned out to have better generalization ability than others.  }
\label{fig:exp-model}
\end{figure*}

\subsection{Effect of Models}
How general are our results across different specific action
recognition models?  To investigate this, we perform experiments
similar to Sec.~\ref{sec:main-exp} with three additional models: (1)
\textbf{3D-ResNet18-Flow}, which keeps the same CNN but feeds in optical
flow frames instead of RGB, (2)
\textbf{SlowFast}~\cite{feichtenhofer2019slowfast}, a variant of two
stream networks which has low frame-rate (slow) and high frame-rate
(fast) pathways fused by lateral connections to model spatial and
temporal information, and (3) \textbf{Spatial Temporal Interaction
  Networks (STIN)}~\cite{materzynska2020something}, a graph
convolution model applied to objects in each video frame, tracking the
hands and objects represented with position and size without any
visual information.

For implementation, we use the same PyTorch default model of
3D-ResNet18, but simply add optical flow as an additional input, and
call it 3D-ResNet18-Flow.
For SlowFast and STIN, use implementations provided by the respective
authors.\footnote{\url{https://github.com/facebookresearch/SlowFast}}
\footnote{\url{https://github.com/joaanna/something_else}}

With the notable exception of STIN, which we will discuss later, we
found that our conclusions about the important properties of training
set structure did not vary across these models: common objects are
very helpful while unique objects are not.  However, we did find
several differences in the accuracies between test types
across models. To illustrate this point, we show the case of a single
common object with 0 to 3 unique objects in Fig~\ref{fig:exp-model}.
ResNet with optical flow has better performance, which makes sense
since it is difficult for it to overfit on object appearance
information. Hence, the accuracy differences between seen
action-object combinations (\ie common and unique-self) and unseen
action-object combinations (\ie unseen and unique-others) are smaller
than the others except for STIN. SlowFast has a similar trend as
ResNet RGB, but it has more gaps between seen action-object
combinations (\ie common and unique-self) and unseen action-object
combinations (\ie unseen and unique-others), suggesting less
generalization ability.

STIN generally has better performance on all four testing types than others. This model is fundamentally different from others: it uses the tracked bounding boxes of hand and object (a single category called \textit{object}, not specific object names) so it has an inductive bias to perform well on unseen objects. However, we should note that this comparison may not be fair to others, since STIN was specifically designed with the Something-Something dataset in mind, while the others were designed for different video classification datasets like Kinetics, and are applicable for a broader range of video classification problems. Moreover, the model relies on an object detector and a tracker to generate inputs; if no objects are detected in the video, the model immediately fails.  We found that the data of the object bounding boxes provided by STIN does not cover all the video clips, and the boxes of a small number of video clips are missing, which may be caused by the failure of object detection or tracking. We had to exclude these video clips without tracked bounding boxes for the STIN experiments, so the test accuracy is not computed on exactly the same sets as others, which may be an unfair comparison if the detection/tracking failures correlated with the difficulty of action recognition.

\section{Discussion and Conclusion}
This paper considers machine learning of action recognition through the lens
of a cognitive mechanism of human learning called cross-situational learning.
We investigated the effect of varying the statistics of
the action-object pairs that make up the training data of action recognition models. More
specifically, we sampled training data from 
action-object co-occurrence matrices having different statistics and compared the resulting
performances of the trained models. We also carefully evaluated the
models with four different types of test action-object statistics.
Our results indicate that having common objects across different
actions is important to learn generalizable action recognition while
having objects unique to actions does not contribute much.
While this finding is intuitive, it is important because most
current mainstream training datasets
for action recognition in computer vision have few common objects across different
actions.

Our study focuses
on the effects of various types of structures in the training data on
the accuracy and generalizability of trained action classifiers. This work is theoretically motivated and complementary to many empirically motivated studies that are
engineering state-of-the-art deep neural networks assuming that the
training data is already given. Instead of simply viewing an action
training dataset as a monolithic set of training examples, we argue
that researchers should view training datasets as having an important
structure that significantly impacts the accuracy of trained
classifiers. Our study provided new insights regarding
 how we should design the training data,  what kind of data
is suitable for training models, and how we should evaluate the
models. Inspired by the cross-situational learning principles, we argued that: (1) training data
should be designed based on the structure of action-object co-occurrences,
(2) designing the dataset with common objects across different actions is
key, and (3) evaluating with four specific types of action-object statistics
yields important insight.
Our study made a first step in examining the structure of the training dataset. We hope it that it will inspire more work that investigates how to design ideal training data for efficiently training
generalizable action recognition models.

\section{Acknowledgements}
This work was supported by the National Science Foundation (CAREER
IIS-1253549, BCS-1842817), the National Institutes of Health (R01 HD074601, R01
HD093792, R01 HD104624), and the U.S. Navy (N00174-19-1-0010), as well as the IU
Office of the Vice Provost for Research, the College of Arts and
Sciences, and the School of Informatics, Computing, and Engineering
through the Emerging Areas of Research Project ``Learning: Brains,
Machines, and Children.''

\bibliography{references}
\bibliographystyle{IEEEtranN}
\clearpage
\onecolumn

\section{Details on neural network training}
 We initialized the network with Kinetics pretrained weights and minimize the standard softmax cross-entropy loss with stochastic gradient descent of Adam with a batch size of 15 and a initial learning rate of $1 \times 10^{-4}$. We decrease the learning rate by a factor of 10 when the validation loss reaches a plateau, and end training when the learning rate reaches $1 \times 10^{-6}$.  The spatiotemporal input size of the model is $112 \times 112 \times 64$, corresponding to width, height, and frame size, respectively. We resize the spatial resolution to $148 \times 112$ via bilinear interpolation and apply random crops for training and center crops for validation and testing.  If the number of frames is less than 64, we pad to 64 by repeating the first and last frames. We use a Nvidia GPU for computation, and the each run finishes less than a day.
 
\section{Details on dense matrix sampling}
We selected the subset co-occurrence matrix of something-something dataset shown in Figure~\ref{fig:matrix-something-selected} by performing the  following heuristics in the order.

\begin{enumerate}
\setlength{\parskip}{0cm} 
\setlength{\itemsep}{0cm} 
    \item Select the objects (columns) that have at least 100 instances in total. 
    \item Select the actions (rows) where 40\% of the cells have at least 15 instances.
    \item Select the objects (columns) where 80\% of the cells have at least 15 instances. 
    \item Remove any row or column that contains a cell with less than 10 instances. 
\end{enumerate}

\section{Details on train/val/test split}
\begin{figure*}[b!]
\begin{center}
	\includegraphics[width=\linewidth]{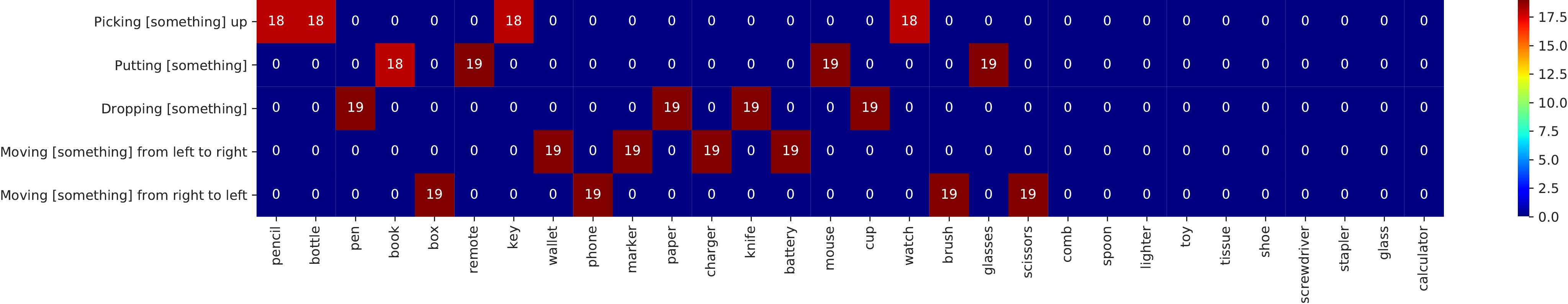}
\end{center}
  \caption{A training matrix sampled for 4 unique objects}
\label{fig:4unique}
\end{figure*}

\begin{figure*}[b!]
\begin{center}
	\includegraphics[width=\linewidth]{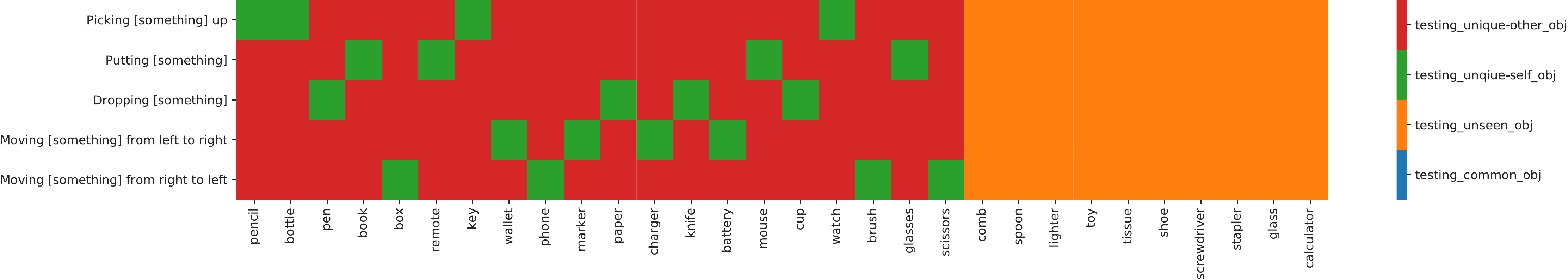}
\end{center}
  \caption{Testing types for the training set of Figure~\ref{fig:4unique}}
\label{fig:4unique-test}
\end{figure*}

\begin{figure*}[b!]
\begin{center}
	\includegraphics[width=\linewidth]{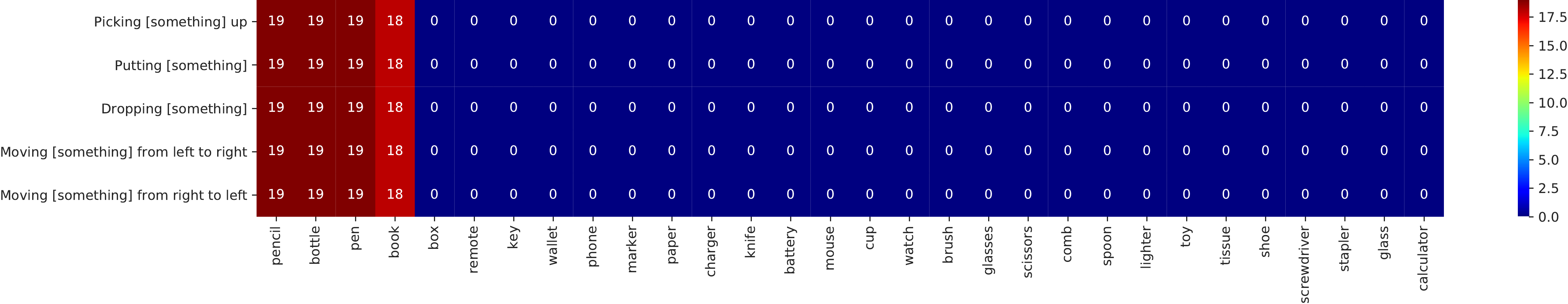}
\end{center}
  \caption{A training matrix sampled for 4 common objects}
\label{fig:4common}
\end{figure*}

\begin{figure*}[b!]
\begin{center}
	\includegraphics[width=\linewidth]{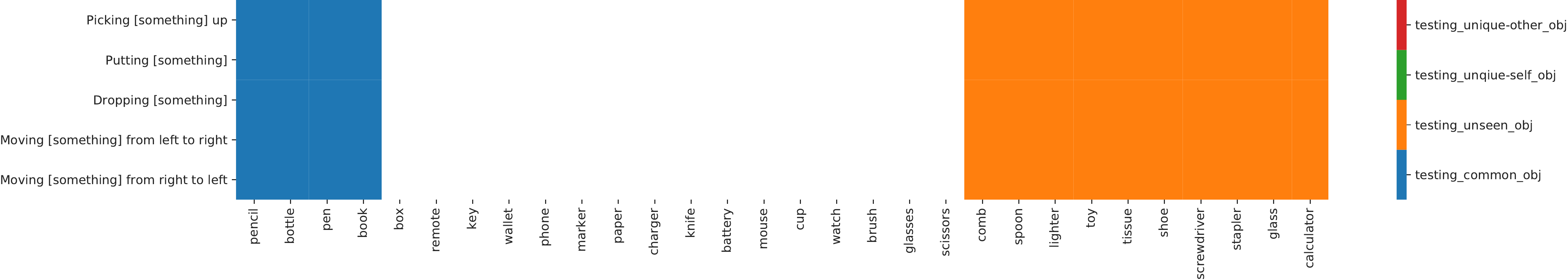}
\end{center}
  \caption{Testing types for the training set of Figure~\ref{fig:4common}}
\label{fig:4common-test}
\end{figure*}

\begin{figure*}
\begin{center}
	\includegraphics[width=\linewidth]{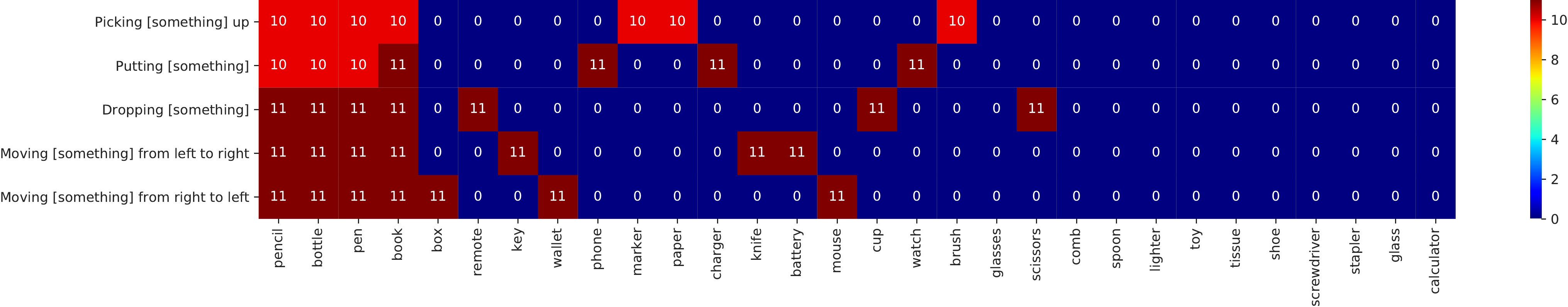}
\end{center}
  \caption{A training matrix sampled for 4 common objects and 3 unique objects}
\label{fig:4common-3unique}
\end{figure*}

\begin{figure*}
\begin{center}
	\includegraphics[width=\linewidth]{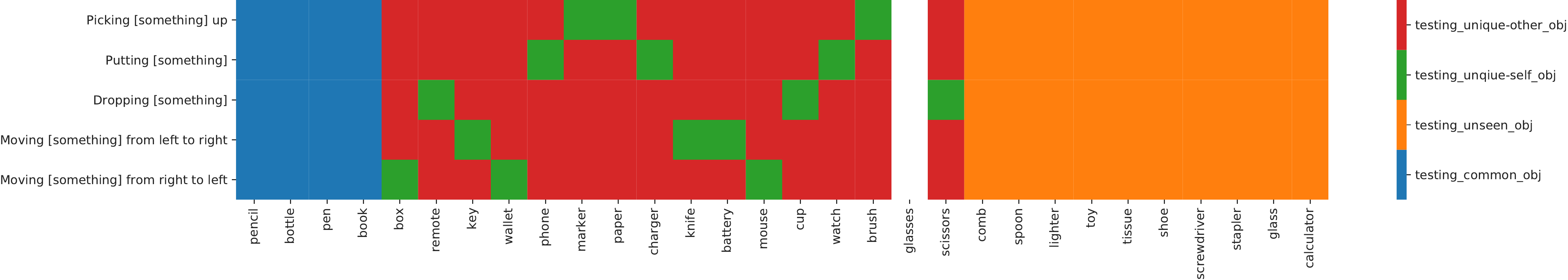}
\end{center}
  \caption{Testing types for the training set of Figure~\ref{fig:4common-3unique}}
\label{fig:4common-3unique-test}
\end{figure*}

Unlike typical studies where train/val/test sets are always fixed, we sample each set every time, repeat the same process multiple (ten in our case) times for each training scenario, and report the mean accuracy with confidence intervals. We basically sample the training instances first, and then use the 80\% of the remaining instances for test, and  20\% for validation. For unseen action-object combinations, we only use the object 21 - 30. We show the examples of 4 unique objects (Figure~\ref{fig:4unique},\ref{fig:4unique-test}), 4 common objects (Figure~\ref{fig:4common},\ref{fig:4common-test}), and 3 unique and 4 common objects (Figure~\ref{fig:4common-3unique},\ref{fig:4common-3unique-test}). 




\end{document}